\documentclass[a4paper, 10pt, conference]{ieeeconf}      
\usepackage{FG2026}
\usepackage{hyperref}
\usepackage{cleveref} 
\usepackage{multirow}
\usepackage[bottom]{footmisc}

\FGfinalcopy 

\IEEEoverridecommandlockouts                              
\def\FGPaperID{266} 

\title{\LARGE \bf
Micro-DualNet: Dual-Path Spatio–Temporal Network for \\Micro-Action Recognition}

\author{\parbox{16cm}{\centering
    {\normalsize Naga VS Raviteja Chappa$^1$, Evangelos Sariyanidi$^1$, Lisa Yankowitz$^1$, Gokul Nair$^1$, \\Casey J. Zampella$^1$$^,$$^2$, Robert T. Schultz$^1$$^,$$^2$ and Birkan Tunç$^1$$^,$$^2$}\\
    {\small
     \{$^1$The Children's Hospital of Philadelphia, USA and $^2$ University of Pennsylvania, USA\}\\
     \small\url{https://compsygroup.github.io/micro-dual-net/}
    \thanks{This work is partially supported by the Office of the Director (OD), National
Institute of Child Health and Human Development (NICHD), and National Institute
of Mental Health (NIMH) of US, under grants R01MH122599, R01MH118327,
P50HD105354 and R21HD102078; and the IDDRC at CHOP/Penn.}
}}}

\usepackage{fancyhdr}
\begin{document}

\ifFGfinal
\thispagestyle{empty}
\pagestyle{empty}
\else
\author{Anonymous FG2026 submission\\ Paper ID \FGPaperID \\}
\pagestyle{plain}
\fi
\maketitle
\thispagestyle{fancy}

\begin{abstract}

Micro-actions are subtle, localized movements lasting 1-3 seconds such as scratching one's head or tapping fingers. Such subtle actions are essential for social communication, ubiquitously used in natural interactions, and thus critical for fine-grained video understanding, yet remain poorly understood by current computer vision systems. We identify a fundamental challenge: micro-actions exhibit diverse spatio-temporal characteristics where some are defined by spatial configurations (e.g., ``covering face") while others manifest through temporal dynamics (e.g., ``leg shaking"). Existing methods that commit to a single spatio-temporal decomposition cannot accommodate this diversity. We propose Micro-DualNet, a dual-path network that processes anatomically-grounded spatial entities through parallel Spatial-Temporal (ST) and Temporal-Spatial (TS) pathways. The ST path captures spatial configurations before modeling temporal dynamics, while the TS path inverts this order to prioritize temporal dynamics. Rather than fixed fusion, we introduce entity-level adaptive routing where each body part learns its optimal processing preference, complemented by Mutual Action Consistency (MAC) loss that enforces cross-path coherence. Extensive experiments demonstrate competitive performance on MA-52 dataset (65.10\% Top-1, 68.72\% F1$_{\text{mean}}$) and state-of-the-art results on iMiGUE (76.88\% Top-1) dataset. Ablations confirm that position-based actions benefit from ST processing while motion-based actions favor TS processing, validating that micro-actions require flexible complementary decomposition. Our work reveals that architectural adaptation to the inherent complexity of micro-actions is essential for advancing fine-grained video understanding. Clinical validation on an in-house dataset of 290 individuals demonstrates that Micro-DualNet-detected micro-actions reveal statistically significant behavioral differences between kids with autism spectrum disorder, other psychiatric conditions, or typical development, suggesting potential for automated behavioral assessment.

\end{abstract}

\section{INTRODUCTION}\label{sec:intro}

Despite their subtlety, the brief movements we unconsciously perform like a head scratch, finger tap, or adjusting glasses, encode substantial behavioral and psychological information. These micro-actions, subtle localized movements lasting 1-3 seconds, are commonly used in natural and spontaneous social interactions, hence critical for social communication. Unlike gross motor actions~\cite{soomro2012ucf101, kuehne2011hmdb, kay2017kinetics} that involve easily discernible full-body motions, micro-actions manifest as brief, small-scale movements of specific body parts. These movements carry significant behavioral and psychological cues critical for applications in behavioral assessment, human-computer interaction, and healthcare monitoring. For instance, subtle differences in motor patterns such as stereotypies may help distinguish autism spectrum disorder (ASD) from other conditions~\cite{goldman2013stereotypies}, yet manual behavioral coding is prohibitively time-intensive for clinical workflows~\cite{grzadzinski2016measuring}. Yet despite recent advances in action recognition~\cite{soomro2012ucf101, kay2017kinetics, wu2021spatiotemporal, luo2021dense, liu2022fineaction}, current methods~\cite{guo2024benchmarking} achieve only 61\% accuracy on micro-action benchmarks, revealing a big challenge for fine-grained video understanding and any application that rely on capturing human behavior from videos.


The core challenge lies in the heterogeneous spatio-temporal characteristics of micro-actions. Consider ``covering face" versus ``stretching arms": the former is characterized by its final spatial configuration while the latter manifests through repetitive temporal patterns where rhythm, not pose or location, carries discriminative information. As illustrated in Fig. 1, this heterogeneity means no single spatio-temporal decomposition captures all micro-actions optimally. Spatial-Temporal (ST) processing, which prioritizes spatial configuration before temporal dynamics, may excel for position-defined actions. Conversely, Temporal-Spatial (TS) processing, which models temporal dynamics before spatial relationships, may better captures motion-defined actions. Current architectures that commit to a single processing order cannot reconcile these opposing requirements.

\begin{figure}[b!]
    \centering
    \includegraphics[width=\linewidth]{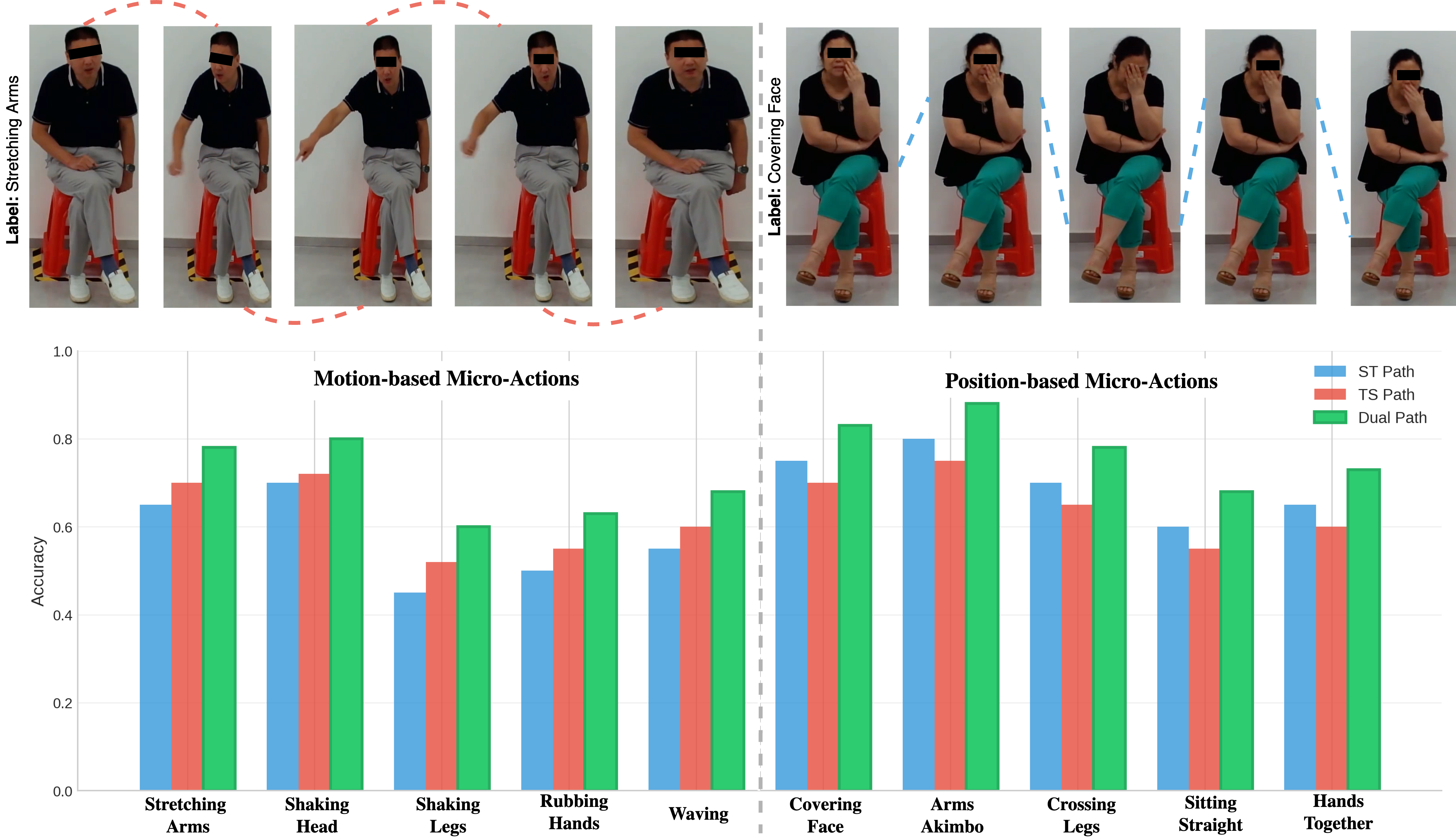}
    \caption{Micro-action recognition requires flexible spatio-temporal processing. (Top) Representative frame sequences showing the distinct characteristics of motion-based vs. position-based micro-actions.``Stretching arms" is defined by repetitive temporal patterns, while ``Covering face" is characterized by its final spatial configuration. (Bottom) Empirical validation on MA-52 dataset: motion-based actions are better modelled by TS path (left), while position-based actions achieve higher accuracy with ST path (right). The dual-path architecture (green) consistently outperforms single paths, demonstrating the necessity of bidirectional processing for comprehensive micro-action understanding. \textbf{Best viewed in color.}}
    \label{fig:motivation}
\end{figure}
Micro-action recognition presents unique challenges beyond traditional action recognition. While traditionally studied actions involve coordinated full-body movements with clearly discernible actions, micro-actions operate within constrained kinematic spaces as they are subtle movements that concentrate discriminative signals in small spatio-temporal regions. These regions shift dynamically with body pose and viewpoint, creating a challenge: the spatial constraint increases recognition difficulty rather than simplifying it. This challenge is manifested in the performance of the current approaches~\cite{guo2024benchmarking, liu2021imigue} that achieve accuracies of 61\% on MA-52 and 71\% on iMiGUE.

The challenge of micro-action recognition is reflected not just in the performance gap, but in the failure modes of existing methods as well i.e., fixed spatial regions misalign under viewpoint changes while single processing orders cannot accommodate both spatially-defined and temporally-defined micro-actions. Analysis of existing methods reveals complementary failure modes that inform our approach. Convolutional Neural Network (CNN)-based methods~\cite{lin2019tsm, wang2018temporal, shao2020temporal} learn appearance features but lack structural priors—when ``touching face" occurs at varying scales or angles, learned spatial filters fail to generalize. Skeleton-based approaches~\cite{cheng2020skeleton, yan2018spatial, chen2021channel, shi2020skeleton} encode anatomical structure but discard appearance cues, losing critical information like hand configurations and surface contacts that distinguish ``rubbing eyes" from ``touching nose." Recent hybrid methods~\cite{duan2022revisiting, gu2025motion} attempt multi-modal fusion but remain committed to a single processing order, missing a key insight: optimal spatio-temporal decomposition varies by action type.

We propose Micro-DualNet, a keypoint-guided dual-path network that adapts to micro-action heterogeneity through complementary spatio-temporal processing. Our approach leverages anatomical keypoints to define six adaptive spatial entities, namely head, face, left hand, right hand, torso, and lower body, via our \emph{spatial entity module (SEM)} [\cref{subsec:sem}], then processes these through parallel \emph{spatial-temporal (ST)} and \emph{temporal-spatial (TS)} pathways [\cref{subsec:modeling}]. The ST path captures spatial entity configurations before modeling temporal dynamics, while the TS path inverts this order to prioritize temporal dynamics. This dual decomposition, combined with a gating/routing mechanism, enables automatic selection of optimal processing strategies per action type.

Rather than fixed fusion, we introduce \emph{entity-level adaptive routing} [\cref{subsubsec:routing}] that allows each body part to learn its optimal blend of ST and TS processing based on its spatio-temporal characteristics. To ensure complementary learning without redundancy, we further propose \emph{Mutual Action Consistency (MAC) loss} [\cref{subsec:macloss}] that enforces cross-path coherence while preserving specialized representations.

We validate our design through extensive experiments on standard benchmarks [\cref{sec:exps}]. Experiments on MA-52~\cite{guo2024benchmarking} and iMiGUE~\cite{liu2021imigue} datasets demonstrate the effectiveness of our approach, with ablations confirming that different body parts benefit from different processing orders. Empirical analysis confirms our hypothesis: position-based actions consistently achieve higher accuracy through ST processing, while motion-defined actions benefit from TS processing. Systematic ablations [\cref{subsec:ablations}] demonstrate that each component contributes meaningfully: Keypoint-guided entities provide robust spatial grounding (+3.8\% over fixed regions), dual paths capture complementary patterns (+9.99\% over single path), and MAC loss ensures effective cooperation (+2.96\%). These results validate that micro-actions indeed require flexible spatio-temporal decomposition, confirming our architectural principles and paving the road for future studies that can achieve higher performance and have more real-life impact. Finally, we provide initial clinical validation demonstrating that Micro-DualNet detected micro-actions reflect meaningful behavioral differences across diagnostic groups, bridging the gap between benchmark performance and real-world utility.

\section{RELATED WORKS}\label{sec: Related-Works}

\noindent\textbf{Micro-action Recognition.}
Micro-actions are subtle, short-duration movements concentrated on specific body parts~\cite{liu2021imigue, chen2023smg}. 
Early methods directly applied standard action recognizers~\cite{liu2024micro, li2023joint} but struggled with the spatially-localized and temporally-brief nature of discriminative signals. 
Guo et al.~\cite{guo2024benchmarking} introduced the MA-52 benchmark and MANet, which combines Temporal Shift Module (TSM)~\cite{lin2019tsm} with spatial entity aggregation using predefined body regions. 
While this establishes useful spatial priors, fixed regions misalign under viewpoint changes.
Recent work addresses these limitations through diverse strategies: 
Motion-Modulated Network (MMN)~\cite{gu2025motion} introduces motion-aware channel modulation for skeleton-based recognition, achieving strong results on MA-52 but lacking appearance information.
MM-Gesture~\cite{gu2025mm} explores multimodal fusion for micro-gestures, while Online Micro-Gesture~\cite{liu2024micro} addresses streaming scenarios.
However, these methods commit to a single processing order (spatial-then-temporal), limiting their ability to capture both configuration-centric and rhythm-centric patterns inherent in micro-actions.

\noindent\textbf{Video CNNs and Transformers.}
Temporal modeling has evolved from sparse sampling (TSN~\cite{wang2018temporal}, TSM~\cite{lin2019tsm}, TS-LSTM~\cite{ma2019ts}) to 3D convolutions (I3D~\cite{carreira2017quo}, SlowFast~\cite{feichtenhofer2019slowfast}) and Video Transformers~\cite{bertasius2021space, liu2022video}. These architectures use global pooling that dilutes localized signals critical for micro-actions. We build on TSM but replace global pooling with keypoint-guided entity extraction and constrain transformer attention to anatomically-grounded regions.

\noindent\textbf{Skeleton-based Methods.}
Graph Convolutional Networks (GCNs)~\cite{yan2018spatial, shi2019two, cheng2020skeleton} and skeleton-specific architectures~\cite{duan2022revisiting} excel when pose estimation is reliable but cannot leverage appearance information.
This limitation proves critical for micro-actions where hand configuration and contact surfaces carry semantic meaning.
While MMN~\cite{gu2025motion} advances skeleton-based micro-action recognition through motion-guided modulation, it cannot distinguish visually-similar actions with different surface interactions.
CTR-GCN~\cite{chen2021channel} improves topology modeling but remains appearance-agnostic.

\noindent\textbf{Part-based Representations.}
Keypoint-guided pooling localizes discriminative regions, with OpenPose~\cite{cao2019openpose} providing reliable body part detection.
MANet~\cite{guo2024benchmarking} employs fixed spatial entities that struggle with viewpoint variations—when a person rotates, predefined regions no longer align with semantic body parts.
Pure pose methods~\cite{gu2025motion, liu2021imigue} sacrifice appearance information entirely.
Our approach addresses both limitations: adaptive keypoint-guided entities maintain semantic alignment across viewpoints while preserving appearance features, and dual-path processing with MAC regularization captures both spatial configurations and temporal dynamics.
Unlike complex ensemble solutions~\cite{lichallenge1, gongchallenge2}, we provide architectural innovations that improve single-model understanding of micro-action structure.

\noindent\textbf{Challenge Solutions and Complex Architectures.}
Recent ACM Multimedia Grand Challenge 2024 submissions~\cite{lichallenge1, gongchallenge2} achieve high accuracy through sophisticated ensembles.
These solutions combine multiple backbones (Swin-L~\cite{liu2022video}, VideoMAE-v2~\cite{wang2023videomae}), extensive data augmentation, and model ensembling.
While demonstrating performance upper bounds, their computational requirements (10-100× our approach) and architectural complexity limit practical deployment.
The winning solution~\cite{lichallenge1} employs five models with test-time augmentation, requiring over 500 GFLOPs per prediction.
In contrast, we focus on architectural insights that improve single-model performance while maintaining efficiency comparable to MANet~\cite{guo2024benchmarking}.





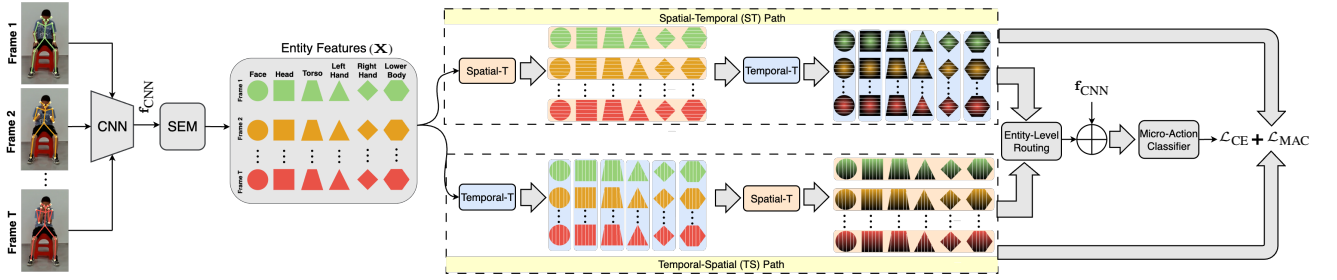
\begin{figure*}[h!]
  \centering
    \includegraphics[width=0.95\linewidth]{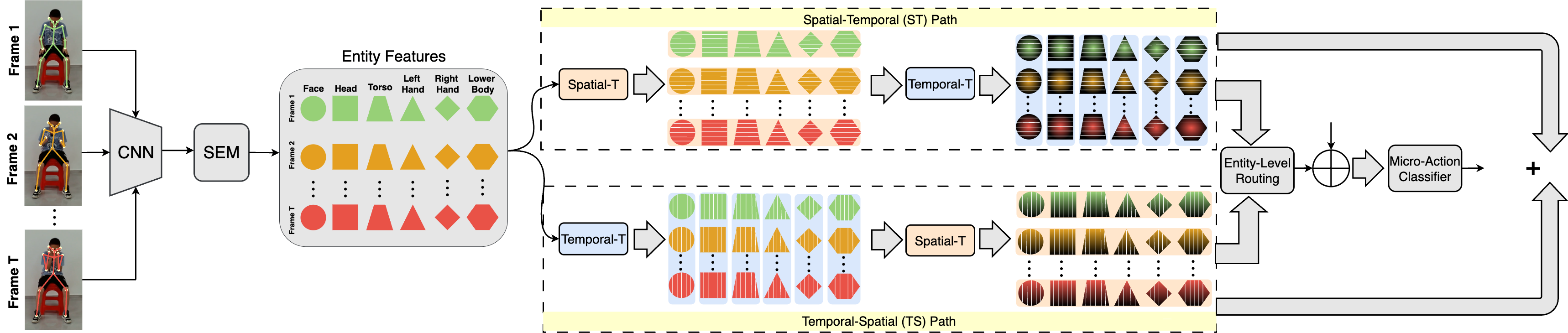}
    \put(-429,60){\rotatebox{90}{\tiny$\mathbf{f}_{\text{CNN}}$}}
    \put(-343,83){{\tiny($\mathbf{X}$)}}
    \put(-79,69){{\tiny$\mathbf{f}_{\text{CNN}}$}}
    \put(-25,50){{\tiny$\mathcal{L}_{\text{CE}}$}}
    \put(-7,50){{\tiny$\mathcal{L}_{\text{MAC}}$}}
   \caption{Overview of the proposed \textbf{Micro-DualNet} framework. Given input video frames with corresponding body joints, our framework extracts CNN features ($\mathbf{f}_{\text{CNN}}$) and decomposes them into anatomically-grounded entity features ($\mathbf{X}$) via the Spatial Entity Module (SEM) [\cref{subsec:sem}]. These entity features are processed through parallel Spatial-Temporal (ST) and Temporal-Spatial (TS) pathways [\cref{subsec:modeling}]: the ST path applies spatial modeling before temporal modeling, while the TS path inverts this order. Entity-level adaptive routing [\cref{subsubsec:routing}] then learns to blend path outputs per body part based on their spatio-temporal characteristics. The routed features are combined with global CNN features and optimized via Cross-Entropy (CE) loss and Mutual Action Consistency (MAC) loss [\cref{subsec:macloss}], which enforces cross-path coherence on raw path outputs before routing. \textbf{Best viewed in color.}}
   \label{fig:framework}
\end{figure*}

\section{METHODOLOGY}\label{sec:methodology}
\subsection{Overview}
As shown in Fig. 2, given an input video $\mathbf{V} \in \mathbb{R}^{T \times H \times W \times 3}$ with $T$ frames, height $H$, width $W$, and 3 color channels, along with corresponding body joints, our framework processes micro-actions through \textbf{four} components: (1) a Spatial Entity Module (SEM) that extracts anatomically-grounded entity representations from Convolutional Neural Network (CNN) features, (2) dual Spatial-Temporal (ST) and Temporal-Spatial (TS) pathways that capture complementary spatio-temporal patterns, (3) entity-level adaptive routing that learns per-entity processing preferences, and (4) Mutual Action Consistency (MAC) loss that enforces cross-path coherence while preserving specialized processing.

\subsection{Spatial Entity Module (SEM)} \label{subsec:sem}
For each frame $t \in \{1,...,T\}$, we extract features $\mathbf{F}_t \in \mathbb{R}^{C \times H' \times W'}$ from the penultimate layer of ResNet-101 with Temporal Shift Module (TSM)~\cite{lin2019tsm}, where $C=2048$, $H'=H/32$, $W'=W/32$. We also use 25 keypoints (joint coordinates) represented as $\{(\mathbf{k}_j, c_j)\}_{j=1}^{25}$ where $\mathbf{k}_j \in \mathbb{R}^2$ denotes pixel coordinates and $c_j \in [0,1]$ indicates detection confidence. Key points can be generated by any human pose detection architecture. In the current study, we used OpenPose~\cite{cao2019openpose} for this purpose. 

We define $K$ anatomical groupings using key points. For MA-52, we use $K=6$ entities (head, face, left\_hand, right\_hand, torso, lower\_body) to capture whole-body micro-actions. For iMiGUE focusing on upper-body micro-gestures, we use $K=5$ entities, excluding lower\_body as the dataset contains seated subjects. Please see Sec. A in Supplementary Material (Supp.) for detailed body joint-to-entity mappings.

Entity bounding boxes are computed dynamically as:
\begin{equation}
\mathbf{B}_{i,t} = \begin{cases}
\text{BBox}(\{\mathbf{k}_j | j \in \mathcal{J}_i, c_j > \theta\}) & \text{if } |\mathcal{V}_i| \geq 2 \\
\mathbf{B}_{i,t-1} & \text{otherwise}
\end{cases}
\end{equation}
where $\mathcal{J}_i$ denotes keypoint indices for entity $i$, $\mathcal{V}_i = \{j \in \mathcal{J}_i | c_j > \theta\}$ are visible keypoints with confidence threshold $\theta=0.3$, and $\text{BBox}(\cdot)$ computes the minimum enclosing rectangle with 10\% padding.

Entity features are extracted via ROIAlign~\cite{he2017mask} followed by entity-specific refinement:
\begin{equation}
\mathbf{x}_{i,t} = \mathcal{E}_i(\text{ROIAlign}(\mathbf{F}_t, \mathbf{B}_{i,t})) + \mathbf{p}_i
\end{equation}
where ROIAlign extracts fixed-size features from arbitrary bounding box regions $\mathbf{B}_{i,t}$ using bilinear interpolation, avoiding the quantization artifacts of ROI pooling; $\mathcal{E}_i$ consists of depthwise separable convolutions projecting to $D=256$ dimensions, and $\mathbf{p}_i \in \mathbb{R}^{D}$ is a learnable position embedding encoding entity identity.

\subsection{Dual-Path Spatio-Temporal Modeling} \label{subsec:modeling}

Given entity features $\mathbf{X} \in \mathbb{R}^{B \times T \times K \times D}$ where $B$ is batch size, $T$ is number of frames, $K$ is number of spatial entities, and $D=256$ is feature dimension, we construct dual paths to capture complementary spatio-temporal patterns in micro-actions.

\subsubsection{Spatial Entity Transformer}
To model spatial relations among body joints within each frame, we design a spatial entity transformer (Spatial-T). For frame $t$, we denote $\mathbf{X}_t \in \mathbb{R}^{B \times K \times D}$ as features of $K$ entities. These features are processed by Spatial-T as:
\begin{align}
\mathbf{X}'_t &= \text{SPE}(\mathbf{X}_t) + \mathbf{X}_t \\
\mathbf{X}''_t &= \text{LN}(\mathbf{X}'_t + \text{MHSA}_s(\mathbf{X}'_t)) \\
\hat{\mathbf{X}}_t &= \text{LN}(\mathbf{X}''_t + \text{FFN}(\mathbf{X}''_t))
\end{align}
where SPE (Spatial Position Encoding) encodes relative spatial positions of entities based on their anatomical hierarchy (head→torso→limbs), MHSA$_s$ (Multi-Head Self-Attention) performs attention across entities to capture inter-entity dependencies crucial for micro-actions, LN denotes Layer Normalization, and FFN (Feed-Forward Network) is a two-layer Multi-Layer Perceptron (MLP) with GELU~\cite{hendrycks2016gaussian} activation. Please see Sec. D in Supp. for more details on the design choices.

\subsubsection{Temporal Transformer}
To capture temporal dynamics of each entity across frames, we employ a temporal transformer (Temporal-T). For entity $i$, we denote $\mathbf{X}_i \in \mathbb{R}^{B \times T \times D}$ as its features across $T$ frames. These features are operated by Temporal-T as:
\begin{align}
\mathbf{X}'_i &= \text{TPE}(\mathbf{X}_i) + \mathbf{X}_i \\
\mathbf{X}''_i &= \text{LN}(\mathbf{X}'_i + \text{MHSA}_t(\mathbf{X}'_i)) \\
\hat{\mathbf{X}}_i &= \text{LN}(\mathbf{X}''_i + \text{FFN}(\mathbf{X}''_i))
\end{align}
where TPE (Temporal Position Encoding) encodes temporal positions using sinusoidal embeddings, and MHSA$_t$ captures motion patterns essential for distinguishing subtle micro-actions.

\subsubsection{Bidirectional Processing Paths}
Micro-actions exhibit diverse spatio-temporal characteristics—some defined by spatial configurations (e.g., ``touch face"), others by temporal patterns (e.g., ``leg shaking"). We arrange transformers in two complementary orders:

\textbf{ST Path:} First captures spatial entity arrangements, then models their temporal evolution:
\begin{align}
\mathbf{X}_{\text{spatial}}^{ST} &= \text{Spatial-T}(\mathbf{X}) + \text{MLP}_s(\mathbf{X}) \\
\mathbf{X}^{ST} &= \text{Temporal-T}(\mathbf{X}_{\text{spatial}}^{ST})
\end{align}
\textbf{TS Path:} First extracts temporal patterns per entity, then models their spatial relationships:
\begin{align}
\mathbf{X}_{\text{temporal}}^{TS} &= \text{Temporal-T}(\mathbf{X}) + \text{MLP}_t(\mathbf{X}) \\
\mathbf{X}^{TS} &= \text{Spatial-T}(\mathbf{X}_{\text{temporal}}^{TS})
\end{align}
The MLPs preserve original entity information through residual connections, enabling adaptive feature combination. Spatial-T and Temporal-T process their respective dimensions efficiently through batch-wise operations: Spatial-T operates on each frame independently (processing $K$ entities per frame), while Temporal-T operates on each entity independently (processing $T$ frames per entity).

\subsubsection{Entity-Level Adaptive Routing} \label{subsubsec:routing}
While the dual-path architecture captures complementary spatio-temporal patterns, a key question remains: how should ST and TS representations be combined? Simple concatenation or addition treats all entities uniformly, ignoring that different anatomical parts exhibit fundamentally distinct characteristics. Hands performing gestures are motion-dominant and benefit from temporal-first processing, while torso postures are configuration-dominant and favor spatial-first processing.

We introduce entity-level adaptive routing that allows each body part to learn its optimal blend of ST and TS representations. For each entity $i \in \{1, ..., K\}$ at each temporal position $t$, given path outputs $\mathbf{X}_{i,t}^{ST}, \mathbf{X}_{i,t}^{TS} \in \mathbb{R}^{D}$, we concatenate them and compute routing scores through a lightweight entity-specific network:
\begin{equation}
\mathbf{r}_{i,t} = \mathcal{R}_i([\mathbf{X}_{i,t}^{ST}; \mathbf{X}_{i,t}^{TS}]) + \mathbf{b}_i
\end{equation}
where $[\cdot;\cdot]$ denotes concatenation, $\mathcal{R}_i: \mathbb{R}^{2D} \rightarrow \mathbb{R}^{2}$ is a two-layer network (Linear-LayerNorm-ReLU-Dropout-Linear), and $\mathbf{b}_i \in \mathbb{R}^{2}$ is a learnable entity-type prior encoding anatomical biases.
The routing scores are converted to normalized weights via temperature-scaled softmax:
\begin{equation}
[\alpha_{i,t}^{ST}, \alpha_{i,t}^{TS}] = \text{softmax}(\mathbf{r}_{i,t} / \tau_r)
\end{equation}
where $\tau_r = 0.7$ controls routing sharpness. The fused entity representation combines both paths according to learned preferences:
\begin{equation}
\mathbf{X}_{i,t}^{\text{fused}} = \alpha_{i,t}^{ST} \cdot \mathbf{X}_{i,t}^{ST} + \alpha_{i,t}^{TS} \cdot \mathbf{X}_{i,t}^{TS}
\end{equation}

\subsection{Mutual Action Consistency Learning}\label{subsec:macloss}

To ensure the dual paths learn consistent representations while maintaining their complementary strengths, we employ entity-aware contrastive learning between the ST and TS pathways.

For each entity $i$ and temporal position $t$, we enforce that representations from both paths align for the same spatio-temporal location while contrasting with different temporal positions:
\begin{equation}
\mathcal{L}_{\text{MAC}}^{i,t} = -\log \frac{\exp(\text{sim}(\mathbf{z}_{i,t}^{ST}, \mathbf{z}_{i,t}^{TS}) / \tau)}{\sum_{j=1}^{T} \exp(\text{sim}(\mathbf{z}_{i,t}^{ST}, \mathbf{z}_{i,j}^{TS}) / \tau)}
\end{equation}
where $\mathbf{z}_{i,t}^{ST}, \mathbf{z}_{i,t}^{TS}$ are $\ell_2$-normalized features for entity $i$ at time $t$ from respective paths, and $\tau=0.07$ is the temperature parameter. This formulation ensures that while both paths process the same entity differently (spatial-first vs temporal-first), they maintain agreement about which temporal segments are most relevant for each entity.
The total MAC loss aggregates across all visible entities and frames:
\begin{equation}
\mathcal{L}_{\text{MAC}} = \frac{\sum_{i=1}^{K} \sum_{t=1}^{T} c_i^t \cdot \mathcal{L}_{\text{MAC}}^{i,t}}{\sum_{i,t} c_i^t}
\end{equation}
where $c_i^t$ represents the keypoint confidence for entity $i$ at time $t$, naturally down-weighting occluded or uncertain entities. Note that MAC loss operates on the raw path outputs $\mathbf{X}^{ST}$ and $\mathbf{X}^{TS}$ \emph{before} adaptive routing (\cref{subsubsec:routing}), providing a training signal that encourages temporal coherence between paths while allowing the routing module to independently learn entity-specific fusion strategies.

\subsection{Training Objectives}\label{subsec:training}

Given the dual-path outputs $\mathbf{X}^{ST}, \mathbf{X}^{TS} \in \mathbb{R}^{B \times T \times K \times D}$, we apply entity-level adaptive routing to obtain fused representations where each entity $i$ is combined according to its learned preferences:
\begin{equation}
\mathbf{X}^{\text{fused}} = \{\mathbf{X}_i^{\text{fused}}\}_{i=1}^{K}
\end{equation}
where $\mathbf{X}_i^{\text{fused}}$ is computed via Eq.~(13)-(15).
The video-level entity representation is obtained by averaging across temporal and entity dimensions:
\begin{equation}
\mathbf{f}_{\text{entity}} = \frac{1}{T \cdot K} \sum_{t=1}^{T} \sum_{i=1}^{K} \mathbf{X}_{i,t}^{\text{fused}}
\end{equation}

This is concatenated with global appearance features for final classification:
\begin{equation}
\mathbf{f}_{\text{final}} = [\mathbf{f}_{\text{CNN}}; \mathbf{f}_{\text{entity}}]
\end{equation}
where $\mathbf{f}_{\text{CNN}} \in \mathbb{R}^{C}$ represents global context obtained via Global Average Pooling (GAP) over the CNN feature maps, capturing scene-level information that complements localized entity features.
The model is trained with classification and consistency objectives:
\begin{equation}
\mathcal{L} = \mathcal{L}_{\text{CE}}(\mathcal{C}(\mathbf{f}_{\text{final}}), y) + \lambda \mathcal{L}_{\text{MAC}}
\end{equation}
where $\mathcal{C}$ is a two-layer MLP classifier, $\mathcal{L}_{\text{CE}}$ is cross-entropy loss, $y$ denotes micro-action labels, and $\lambda=0.1$ balances the objectives. Importantly, $\mathcal{L}_{\text{MAC}}$ is computed on raw path outputs before routing, ensuring both paths receive gradient signals regardless of learned routing preferences. This design separates representation learning (via MAC) from adaptive fusion (via routing), allowing each component to fulfill its distinct role.

\section{EXPERIMENTAL RESULTS}\label{sec:exps}
\begin{table*}[ht]
\caption{Performance comparison on MA-52~\cite{guo2024benchmarking} and iMiGUE~\cite{liu2021imigue} datasets.}
\label{tab:dual_dataset_results}
\centering
\footnotesize
\renewcommand{\arraystretch}{1.05}
\setlength{\tabcolsep}{2.5pt}
\begin{tabular}{lcccccccccc}
\toprule
\multirow{4}{*}{\textbf{Method}} & \multicolumn{8}{c}{\textbf{MA-52}} & \multicolumn{2}{c}{\textbf{iMiGUE}} \\
\cmidrule(lr){2-9} \cmidrule(lr){10-11}
& \multicolumn{3}{c}{\textit{Accuracy (\%)}} & \multicolumn{4}{c}{\textit{F1 Score (\%)}} & \textit{F1$_{\text{mean}}$} & \multicolumn{2}{c}{Accuracy (\%)} \\
\cmidrule(lr){2-4} \cmidrule(lr){5-8} \cmidrule(lr){9-9} \cmidrule(lr){10-11}
& Body & Action & Action & \multicolumn{2}{c}{Body} & \multicolumn{2}{c}{Action} & Overall & Top-1 & Top-5 \\
\cmidrule(lr){2-2} \cmidrule(lr){3-3} \cmidrule(lr){4-4}\cmidrule(lr){5-6} \cmidrule(lr){7-8}
& Top-1 & Top-1 & Top-5 & Macro & Micro & Macro & Micro &  & & \\
\midrule
TSN~\cite{wang2018temporal} & 59.22 & 34.46 & 73.34 & 52.50 & 59.22 & 28.52 & 34.46 & 43.67 & 51.54 & 85.42 \\
TIN~\cite{shao2020temporal} & 73.26 & 52.81 & 85.37 & 66.99 & 73.26 & 39.82 & 52.81 & 58.22 & 52.38 & 86.15 \\
TSM~\cite{lin2019tsm} & 77.64 & 56.75 & 87.47 & 70.98 & 77.64 & 40.19 & 56.75 & 61.39 & 61.10 & 91.24 \\
MANet~\cite{guo2024benchmarking} & 78.95 & 61.33 & 88.83 & 72.87 & 78.95 & 49.22 & 61.33 & 65.59 & 62.54 & 92.18 \\
\hline
C3D~\cite{tran2015learning} & 74.04 & 52.22 & 86.97 & 66.60 & 74.04 & 40.86 & 52.22 & 58.43 & 20.32 & 55.31 \\
I3D~\cite{carreira2017quo} & 78.16 & 57.07 & 88.67 & 71.56 & 78.16 & 39.84 & 57.07 & 61.66 & 34.96 & 63.69 \\
SlowFast~\cite{feichtenhofer2019slowfast} & 77.18 & 59.60 & 88.54 & 70.61 & 77.18 & 44.96 & 59.60 & 63.09 & 58.73 & 89.41 \\
\hline
VideoSwin-T~\cite{liu2022video} & 77.95 & 57.23 & 87.99 & 71.25 & 77.95 & 38.53 & 57.23 & 61.24 & 55.82 & 88.67 \\
TimesFormer~\cite{bertasius2021space} & 69.17 & 40.67 & 82.67 & 61.90 & 69.17 & 34.38 & 40.67 & 51.53 & 48.15 & 82.34 \\
UniFormer~\cite{li2022uniformer} & 79.03 & 58.89 & 87.29 & 71.80 & 79.03 & 48.01 & 58.89 & 64.43 & 57.29 & 89.95 \\
\hline
ST-GCN~\cite{yan2018spatial} & 69.87 & 49.61 & 79.54 & 61.53 & 69.87 & 34.64 & 49.61 & 53.91 & 46.97 & 84.09 \\
2s-AGCN~\cite{shi2019two} & 70.07 & 49.48 & 78.27 & 61.30 & 70.07 & 34.64 & 49.48 & 53.87 & 47.78 & 88.43 \\
Shift-GCN~\cite{cheng2020skeleton} & 71.23 & 51.85 & 80.16 & 62.48 & 71.23 & 36.92 & 51.85 & 55.62 & 51.51 & 88.18 \\
CTR-GCN~\cite{chen2021channel} & 72.06 & 52.61 & 81.22 & 63.46 & 72.06 & 37.79 & 52.61 & 56.48 & 52.94 & 89.76 \\
\hline
PoseConv3D~\cite{duan2022revisiting} & 80.95 & 63.52 & 90.23 & 74.96 & 80.95 & 47.20 & 63.52 & 66.66 & 64.38 & 93.52 \\
PCAN~\cite{li2025prototypical} & 82.30 & \textbf{66.74} & 91.75 & 77.02 & 82.30 & 53.83 & \textbf{66.74} & \textbf{69.97} & -- & -- \\
\hline
Ours (Pose Only) & 79.64 & 61.25 & 89.42 & 73.18 & 79.64 & 46.73 & 61.25 & 65.20 & 68.92 & 94.35 \\
Ours (RGB Only) & 81.18 & 62.87 & 90.68 & 75.42 & 81.18 & 48.95 & 62.87 & 67.11 & 71.54 & 95.18 \\
\textbf{Ours (Pose + RGB)} & \textbf{83.50} & {65.10} & \textbf{92.27} & \textbf{78.31} & \textbf{83.50} & \textbf{54.18} & {65.10} & {68.72} & \textbf{76.88} & \textbf{96.72} \\
\bottomrule
\end{tabular}
\vspace{-2mm}
\end{table*}

\subsection{Datasets and Evaluation Metrics}
\noindent\textbf{MA-52 Dataset}~\cite{guo2024benchmarking} is a large-scale micro-action dataset collected through psychological interviews capturing unconscious human micro-behaviors. The dataset contains 22,422 samples annotated hierarchically at two levels: 7 body-level and 52 action-level categories. Following standard splits defined in~\cite{guo2024benchmarking}, we use 11,250, 5,586, and 5,586 samples for training, validation, and testing respectively. The dataset provides both RGB frames and OpenPose body joints, enabling multi-modal analysis. Actions span 1-3 seconds and include subtle movements like ``touching face," ``leg shaking," and ``arms crossing".

\noindent\textbf{iMiGUE Dataset}~\cite{liu2021imigue} focuses on upper-body micro-gestures collected from sports interviews. The dataset contains 32 micro-gesture categories with 12,899, 777, and 4,562 samples for training, validation, and testing respectively. While conceptually similar to micro-actions, these micro-gestures are restricted to upper limbs, making skeleton data particularly relevant. We evaluate on iMiGUE to demonstrate our method's generalization beyond full-body movements.

\noindent\textbf{Evaluation Metrics.}
Following standard practice in micro-action recognition~\cite{guo2024benchmarking, liu2021imigue}, we adopt Top-1/Top-5 accuracy, micro and macro F1 scores as evaluation metrics. While accuracy provides direct classification performance, F1 score better handles class imbalance inherent in micro-action datasets. We compute F1$_{\text{mean}}$ by averaging macro and micro F1 scores across both hierarchical levels (body-level and action-level), providing a balanced assessment across different granularities and class frequencies.
\subsection{Implementation Details}
\textbf{Architecture Configuration.} We employ ResNet-101 with TSM \cite{lin2019tsm} as our backbone, pretrained on Kinetics-400~\cite{kay2017kinetics}. SEM extracts $K=6$ entities for MA-52 and $K=5$ entities for iMiGUE datasets with dimension $D=256$ each. Both ST and TS paths use 3-layer transformers with 8 attention heads, hidden dimension 1024, and dropout 0.1. For entity-level adaptive routing, each of the $K$ entities has a dedicated routing network $\mathcal{R}_i$ consisting of two linear layers ($2D \rightarrow D/2 \rightarrow 2$, i.e., $512 \rightarrow 128 \rightarrow 2$) with LayerNorm, ReLU activation, and dropout 0.1. The learnable entity-type bias $\mathbf{b}_i$ is initialized to zero. The routing temperature $\tau=1.0$ provides soft routing that allows continuous blending between paths. This module adds approximately 0.5M parameters ($\sim$3\% overhead). The final classifier uses a 2-layer MLP (512→256→$C$) with GeLU~\cite{hendrycks2016gaussian} activation and dropout 0.5, where $C$ is the number of action classes (52 for MA-52, 32 for iMiGUE as shown in Fig. 4). For MAC loss computation, we use temperature $\tau=0.07$ and $\lambda_{MAC}=0.1$.

\noindent\textbf{Training Details.} We sample 8 frames with temporal stride 8 for MA-52 dataset and 16 frames with stride 4 for iMiGUE dataset. Data augmentation includes random cropping to 224×224, horizontal flipping (p=0.5), and temporal jittering. We train with SGD~\cite{ruder2016overview} optimizer (momentum 0.9), initial learning rate 0.01 with cosine annealing schedule, batch size of 12, and weight decay of $5 \times 10^{-4}$ for 120 epochs. The first 10 epochs use linear warmup from 0.001 to 0.01. Implementation was done on a single GPU (Nvidia RTX 3090).

\subsection{Comparison with State-of-the-Art Methods}

Table~\ref{tab:dual_dataset_results} compares our method with state-of-the-art approaches. On MA-52, we achieve 68.72\% F1$_{\text{mean}}$, competitive with PCAN~\cite{li2025prototypical} (69.97\%) while using simple end-to-end training instead of PCAN's complex 3-stage pipeline. On iMiGUE, we achieve state-of-the-art 76.88\% Top-1 accuracy, surpassing PoseConv3D~\cite{duan2022revisiting} by 12.50\%.

Our adaptive keypoint-guided entities outperform MANet's fixed regions~\cite{guo2024benchmarking} by 3.13\% F1$_{\text{mean}}$. While 3D CNNs (SlowFast~\cite{feichtenhofer2019slowfast}: 63.09\%) and Transformers (UniFormer~\cite{li2022uniformer}: 64.43\%) achieve reasonable performance, they require substantially more computation. Pure skeleton methods struggle—CTR-GCN~\cite{chen2021channel} achieves only 56.48\%, confirming that appearance cues are indispensable for micro-actions.

The bottom rows of~\cref{tab:dual_dataset_results} show modality ablations: pose-only (65.20\%) surpasses all skeleton baselines, RGB-only (67.11\%) demonstrates entity-aware processing benefits, and their fusion (68.72\%) confirms complementarity. Even single-modality variants outperform heavier baselines, validating our dual-path design and MAC regularization.
The 12.50\% improvement on iMiGUE versus 3.13\% on MA-52 reveals that adaptive entity extraction particularly excels for concentrated upper-body micro-gestures, where our keypoint guidance maintains semantic alignment across viewpoints while fixed regions fail.

\subsection{Ablation Studies} \label{subsec:ablations}

\noindent\textbf{Contribution of each component.} 
Table~\ref{tab:ablation_compact} presents systematic analysis of each component's contribution. Starting from the TSM baseline (52.15\% on MA-52), adding a single TS path improves to 55.96\% (+3.81\%) by capturing temporal dynamics. The Spatial Entity Module provides substantial gains (+3.25\% MA-52, +5.39\% iMiGUE), confirming that keypoint-guided entity extraction outperforms global features. Adding the ST path for dual-path processing further improves to 62.14\% (+2.93\%), validating that ST and TS paths capture complementary patterns. Analyzing the fusion components separately: MAC loss alone provides +2.26\% on MA-52 and +2.87\% on iMiGUE by enforcing cross-path temporal coherence, while entity routing alone contributes +0.95\% and +0.83\% respectively. Crucially, combining both achieves +2.96\% and +4.23\%, exceeding MAC alone on iMiGUE, demonstrating synergistic effects where MAC ensures well-formed path representations while routing learns optimal entity-specific combinations. Overall, our full model improves 12.95\% on MA-52 and 18.15\% on iMiGUE over the baseline.
\begin{table}[t]
\centering
\caption{Ablation study of Micro-DualNet components on MA-52 and iMiGUE datasets.}
\label{tab:ablation_compact}
\setlength{\tabcolsep}{3pt}
\footnotesize
\begin{tabular}{@{}lcccccc@{}}
\toprule
\multirow{2}{*}{\textbf{Configuration}} & \multicolumn{4}{c}{\textbf{Components}} & \multicolumn{2}{c}{\textbf{Top-1 (\%)}} \\
\cmidrule(lr){2-5} \cmidrule(lr){6-7}
& SEM & Dual-path & MAC & Routing & MA-52 & iMiGUE \\
\midrule
Baseline (TSM) & & & & & 52.15 & 58.73 \\
+ TS Only & & & & & 55.96 & 63.48 \\
+ SEM & \checkmark & & & & 59.21 & 68.87 \\
+ Dual Path & \checkmark & \checkmark & & & 62.14 & 72.65 \\
+ MAC Loss & \checkmark & \checkmark & \checkmark & & 64.40 & 75.52 \\
+ Routing - MAC & \checkmark & \checkmark & & \checkmark & 63.09 & 73.48 \\
\textbf{Full Model} & \checkmark & \checkmark & \checkmark & \checkmark & \textbf{65.10} & \textbf{76.88} \\
\bottomrule
\end{tabular}
\end{table}

\noindent\textbf{Impact of Spatial Entity Extraction Methods.}
Table~\ref{tab:spatial_regions} compares different spatial entity extraction strategies. Center crop (58.3\% Acc.) extracts only the central region of each frame, losing peripheral information crucial for micro-actions involving limbs or off-center movements. Fixed body regions, similar to MANet~\cite{guo2024benchmarking}, achieve 61.3\% accuracy but fail under pose variations. 
Our keypoint-guided approach achieves 65.1\% accuracy by dynamically adapting entity boundaries based on detected joints. Removing confidence weighting drops performance to 63.1\%, as unreliable keypoints corrupt entity features. Using only 4 entities (excluding lower body) reduces accuracy to 61.4\%, confirming the importance of full-body modeling for MA-52. The 3.8\% improvement over fixed regions validates adaptive entity extraction's importance for handling pose variations and subtle movements.
\begin{table}[t]
\centering
\caption{Ablation study of spatial entity extraction methods on MA-52~\cite{guo2024benchmarking} dataset.}
\label{tab:spatial_regions}
\begin{tabular}{lcc}
\toprule
\textbf{Method} & \textbf{Top-1 (\%)} & \textbf{F1-mean} \\
\midrule
Center Crop & 58.3 & 0.594 \\
Fixed Body Regions & 61.3 & 0.656 \\
\midrule
\textbf{Keypoint-Guided (Ours)} & \textbf{65.1} & \textbf{0.687} \\
\quad w/o confidence weighting & 63.1 & 0.657 \\
\quad w/ 4 entities (no lower body) & 61.4 & 0.641 \\
\bottomrule
\end{tabular}
\end{table}

\noindent\textbf{Impact of temporal modeling.}
Table~\ref{tab:temporal_ablation} investigates temporal design choices. Frame sampling shows gains from 4 frames (59.8\% Top-1) to 16 frames (65.1\%), while 32 frames (64.7\%) slightly degrades, suggesting temporal redundancy beyond 16 frames for 1-3 second micro-actions.
For temporal aggregation, simple pooling (avg: 65.7\%, max: 61.4\%) discards temporal ordering. LSTM (62.8\%) and Temporal Transformer (63.5\%) preserve structure but process entities independently, missing critical inter-entity coordination. Our dual-path design (65.1\%) captures synchronized movements essential for actions like ``rubbing hands" where hand coordination defines the action.
MAC loss granularity affects consistency: frame-level (62.7\%) over-constrains at every timestamp, while video-level (65.1\%) optimally balances by enforcing entity-wise consistency with temporal flexibility. Please see Supp. for additional ablation experiments.

\begin{table}[b]
\centering
\caption{Impact of temporal modeling and frame sampling on MA-52~\cite{guo2024benchmarking} dataset.}
\label{tab:temporal_ablation}
\begin{tabular}{lcc}
\toprule
\textbf{Method} & \textbf{Top-1 (\%)} & \textbf{F1-mean} \\
\midrule
\multicolumn{3}{c}{\textit{Frame Sampling}} \\
4 frames & 59.8 & 0.638 \\
8 frames & 63.2 & 0.663 \\
16 frames & \textbf{65.1} & \textbf{0.687} \\
32 frames & 64.7 & 0.672 \\
\midrule
\multicolumn{3}{c}{\textit{Temporal Aggregation}} \\
Average Pooling & 61.4 & 0.657 \\
Max Pooling & 60.2 & 0.641 \\
LSTM & 62.8 & 0.659 \\
Temporal Transformer & 63.5 & 0.668 \\
\textbf{(Ours)} & \textbf{65.1} & \textbf{0.687} \\
\midrule
\multicolumn{3}{c}{\textit{Entity Temporal Granularity}} \\
Frame-level MAC & 62.7 & 0.657 \\
\textbf{Video-level MAC (Ours)} & \textbf{65.1} & \textbf{0.687} \\
\bottomrule
\end{tabular}
\end{table}

\subsection{Clinical Validation Study} \label{subsec:clinical}
To evaluate the potential clinical utility, we applied Micro-DualNet to an in-house dataset of 290 individuals (ages 5--52) recorded during 2--3 minute conversations with a research staff member [The study which includes this dataset was reviewed and approved by the Institutional Review Board at CHOP.]. Participants received licensed psychologist-supervised diagnostic evaluations and were classified into three groups: ASD (autism spectrum disorder, $n$=120), PSY (non-autistic psychiatric conditions, $n$=46), and TDC (typically developing, $n$=124).
For the ten most frequent micro-actions, we conducted pairwise group comparisons using two-part (hurdle) analysis: (a) probability of engagement via logistic GLM (Prob.), and (b) intensity among engagers via fractional logit (Int.). Table~\ref{tab:clinical} summarizes significant group differences. Notably, the PSY group showed elevated ``retracting feet'' intensity compared to both ASD ($p < 0.001$) and TDC ($p < 0.01$), while ``turning head'' intensity was lower in PSY than both ASD and TDC ($p < 0.05$). Fig. 3 illustrates intensity distributions for these two micro-actions; see Supp. for all comparisons. Although analyses controlling for demographics are needed for definitive interpretation, these results provide initial evidence that micro-action detection can be used to identify behavioral differences in psychiatric conditions as part of a larger computational behavior analysis research.

\begin{figure}[t]
    \centering
    \includegraphics[width=0.6\linewidth]{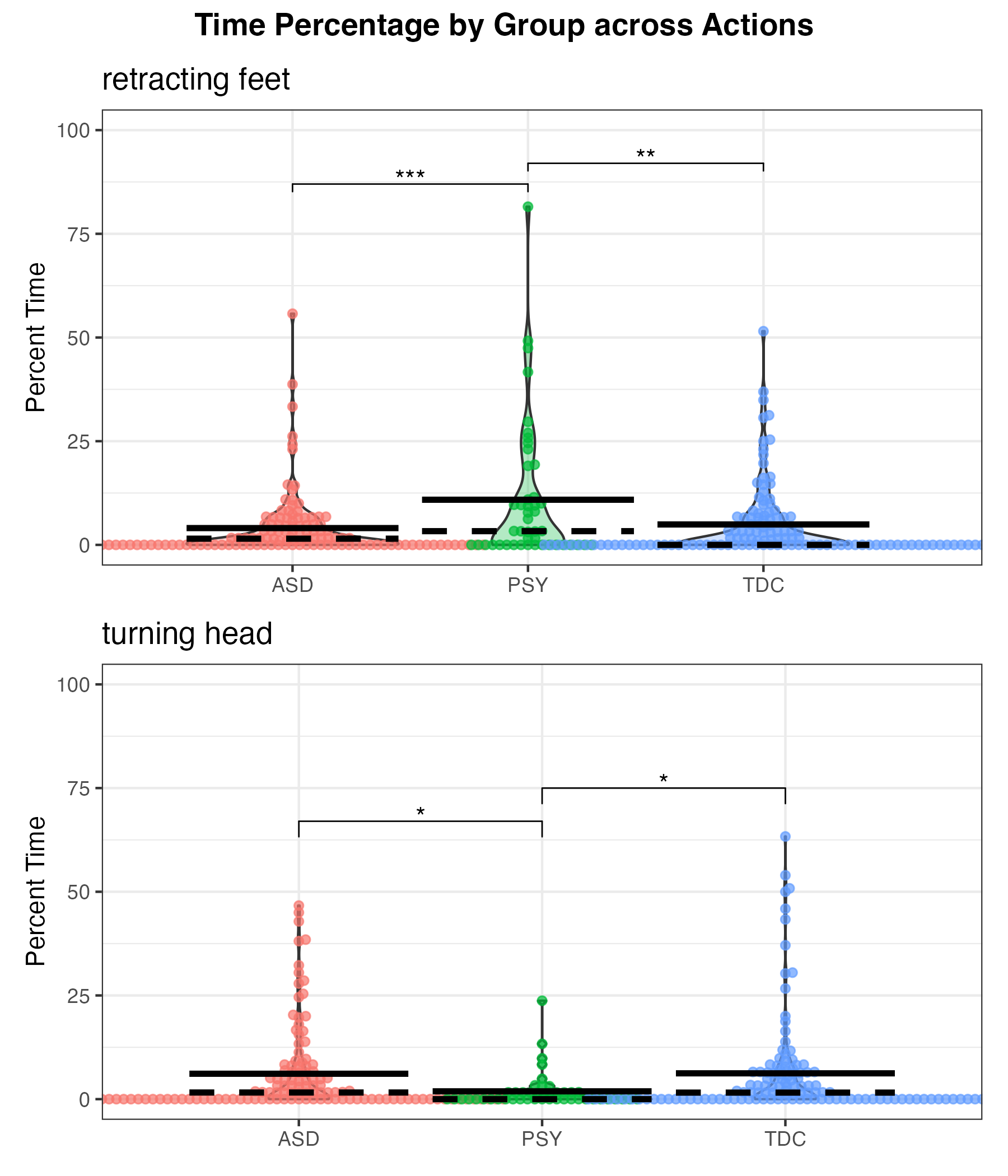}
    \caption{Violin plots of percent time engaged in ``retracting feet'' and ``turning head'' by diagnostic group. Each dot represents one participant's mean percent time engaged. Dashed lines indicate group medians; solid lines indicate group means. Brackets with asterisks denote statistically significant between-group differences in intensity among participants with nonzero engagement: $^{*}p < 0.05$, $^{**}p < 0.01$, $^{***}p < 0.001$. PSY shows significantly elevated ``retracting feet'' intensity compared to both ASD and TDC. }
    \label{fig:clinical}
\end{figure}
\begin{table}[t]
\centering
\caption{Clinical validation: pairwise group differences ($p < 0.05$) in micro-action engagement. Prob.: engagement probability; \%: Percentage of time engaged among participants with nonzero engagement. $p_{\text{adj}}$: adjusted $p$-value correcting for multiple comparisons.}
\label{tab:clinical}
\setlength{\tabcolsep}{2pt}
\scriptsize
\begin{tabular}{@{}llcccc@{}}
\toprule
\textbf{Action} & \textbf{Contrast} & \textbf{Type} & \textbf{Effect} & \textbf{$p$} & \textbf{$p_{\text{adj}}$} \\
\midrule
retracting feet & ASD $<$ PSY & \% & 0.37 & $<$.001 & \textbf{.004} \\
shaking legs & ASD $>$ PSY & Prob. & 3.00 & .002 & .054 \\
shaking head & ASD $<$ PSY & Prob. & 0.27 & .004 & .054 \\
retracting feet & PSY $>$ TDC & \% & 1.96 & .007 & .106 \\
head up & ASD $>$ TDC & Prob. & 2.37 & .007 & .058 \\
stretching feet & ASD $<$ TDC & Prob. & 0.50 & .008 & .058 \\
stretching feet & ASD $<$ PSY & \% & 0.48 & .017 & .172 \\
tilting head & ASD $>$ PSY & Prob. & 2.31 & .019 & .102 \\
nodding & ASD $<$ PSY & Prob. & 0.23 & .020 & .102 \\
turning head & ASD $>$ PSY & \% & 2.71 & .036 & .251 \\
\bottomrule
\end{tabular}
\end{table}
\begin{figure*}[h!]
    \centering
    \includegraphics[width=0.80\linewidth]{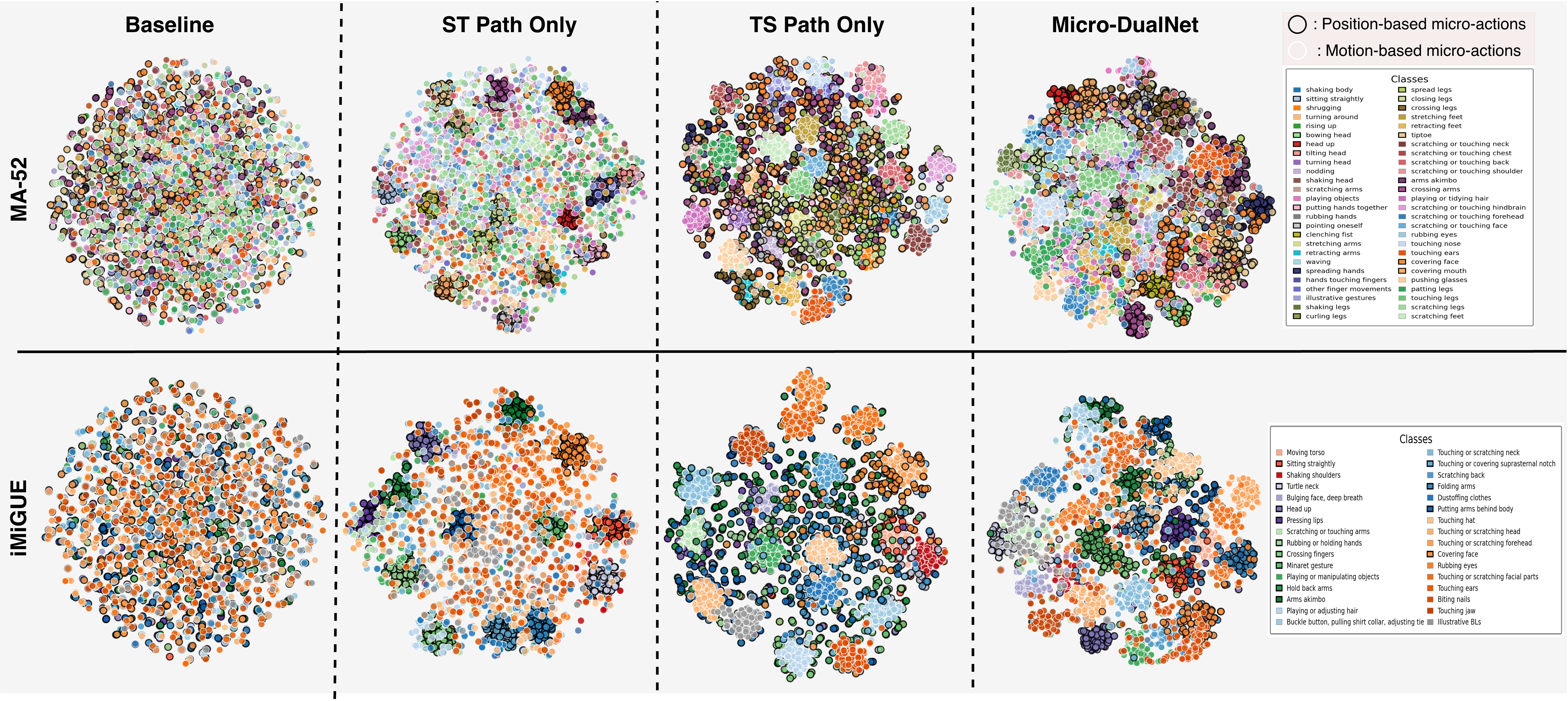}
    \caption{\textbf{t-SNE}~\cite{van2008visualizing} \textbf{visualizations} of learned representations on MA-52~\cite{guo2024benchmarking} (top) and iMiGUE~\cite{liu2021imigue} (bottom) datasets. Baseline (left) shows heavily overlapping classes, single ST/TS paths (middle) yield partial clustering improvements, while Micro-DualNet (right) shows improved, though not complete, class grouping; with several action categories forming more coherent clusters. The remaining overlap reflects the inherent difficulty of fine-grained micro-action discrimination. \textbf{Best viewed in zoom and color.}}
    \label{fig:tsne}
\end{figure*}

\begin{figure*}[ht!]
    \centering
    \includegraphics[width=0.85\linewidth]{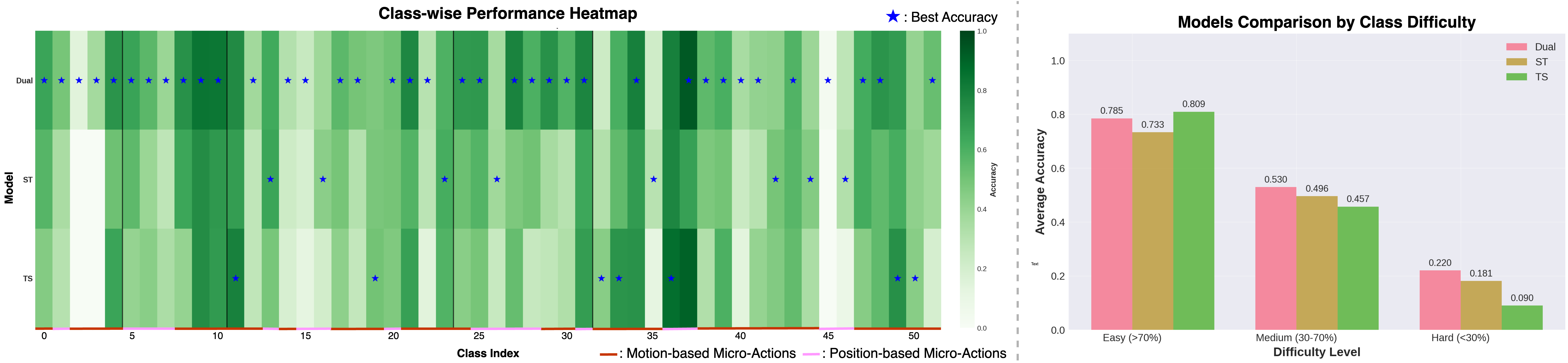}
    \caption{(Left) Class-wise accuracy heatmap for Dual, ST, and TS paths across MA-52~\cite{guo2024benchmarking} dataset. (Right) Model performance by class difficulty: dual-path achieves largest gains on hard categories (+31\%), validating its effectiveness for ambiguous actions. \textbf{Best viewed in zoom and color.}}
    \label{fig:performance_comparison}
\end{figure*}

\subsection{Qualitative Results} \label{sec:qual-results}
t-SNE visualizations (Fig. 4) show that single-path models yield complementary patterns—ST groups position-based actions while TS separates motion-based ones and Micro-DualNet combines these strengths with improved overall clustering. This aligns with Fig. 5: Micro-DualNet shows modest 3\% gains on easy actions but 31\% improvement on hard actions, suggesting that complementary entity-centric processing most benefits challenging micro-actions where single paths struggle.

\section{DISCUSSION}\label{sec:Discussion}
Our results reveal key insights. First, keypoint-guided entities show improved performance over fixed regions in our experiments, though further validation is needed. Second, contrasting ST/TS performance patterns—ST excelling on position-defined actions, TS on motion-based ones—validate that micro-actions require flexible processing. Third, larger gains on iMiGUE (12.5\%) versus MA-52 (3.1\%) suggest our approach particularly benefits concentrated micro-gestures.
\noindent\textbf{Clinical Implications.} Automatically-detected micro-actions differ significantly across diagnostic groups. Elevated ``retracting feet'' in PSY and increased ``shaking legs'' in ASD align with established phenotypes~\cite{nuckols2013diagnostic, leekam2011restricted}. These findings suggest Micro-DualNet could support scalable behavioral assessment.
\vspace{-3mm}
\section{CONCLUSIONS}
We presented Micro-DualNet, a keypoint-guided dual-path framework for micro-action recognition. By processing anatomically-grounded entities through parallel ST and TS pathways with entity-level adaptive routing and MAC regularization, we achieve competitive performance on commonly used datasets. Beyond benchmarks, clinical validation demonstrates that detected micro-actions may reveal significant behavioral differences across ASD, psychiatric, and typically developing groups, providing initial evidence for real-world clinical utility. Our key contribution is demonstrating that micro-actions require flexible entity-level spatio-temporal processing, combined with interpretable routing that could inform automated behavioral assessment in healthcare settings.
\noindent\textbf{Limitations.} 
Our method depends on an external keypoint detector, making it vulnerable to pose estimation failures under severe occlusions. The dual-path architecture increases cost ($\sim$1.9$\times$ single path), and learned routing patterns may not transfer across datasets without fine-tuning. Fixed entity definitions may not optimally capture all micro-action types. Our clinical validation requires confirmation with demographic controls. Future work should explore learnable entity discovery, cross-dataset transfer, and expanded clinical evaluation.






{\small
\bibliographystyle{ieee}
\bibliography{egbib}
}

\end{document}